\documentclass{article}
\usepackage{ijcai26}

\usepackage{times}
\usepackage{soul}
\usepackage{url}
\usepackage[hidelinks]{hyperref}
\usepackage[utf8]{inputenc}
\usepackage[small]{caption}
\usepackage{algorithm}
\usepackage{algorithmic}
\usepackage[switch]{lineno}
\usepackage{natbib}
\usepackage{subcaption}
\newcommand{\HalluVerse}{HalluVerseM$^{3}$}
\usepackage{amsmath, amssymb, amsfonts}
\usepackage{mathtools}
\usepackage{xcolor}
\usepackage{microtype}
\usepackage{pifont}

\usepackage{booktabs}
\usepackage{multirow}

\PassOptionsToPackage{table}{xcolor}
\usepackage{colortbl}
\newcommand{\hmA}[1]{\cellcolor{blue!6}\strut #1}
\newcommand{\hmB}[1]{\cellcolor{blue!14}\strut #1}
\newcommand{\hmC}[1]{\cellcolor{blue!26}\strut #1}
\newcommand{\hmD}[1]{\cellcolor{blue!40}\strut #1}
\newcommand{\hmE}[1]{\cellcolor{blue!56}\strut #1}
\newcommand{\hmF}[1]{\cellcolor{blue!72}\strut #1}
\newcommand{\best}[1]{\textbf{#1}}
\newcommand{\legbox}[2]{%
  \colorbox{#1}{\tiny\hspace{1pt}#2\hspace{1pt}}%
}
\usepackage{tikz}
\usepackage{graphicx}   
\newcommand{\kval}[1]{\textcolor{black!80}{#1}}

\title{Halluverse-M\textsuperscript{3}: A multitask multilingual benchmark for hallucination in LLMs}
\author{
Samir Abdaljalil$^1$
\and
Parichit Sharma$^2$\and
Erchin Serpedin$^{1}$\And
Hasan Kurban$^3$\\
\affiliations
$^1$Texas A\&M University, College Station, TX USA\\
$^2$University of Maryland, Baltimore\\
$^3$Hamad Bin Khalifa University, Doha, Qatar\\
\emails
sabdaljalil@tamu.edu, hkurban@hbku.edu.qa
}
\begin{document}
\maketitle

\begin{abstract}
    Hallucinations in large language models remain a persistent challenge, particularly in multilingual and generative settings where factual consistency is difficult to maintain. While recent models show strong performance on English-centric benchmarks, their behavior across languages, tasks, and hallucination types is not yet well understood. In this work, we introduce \HalluVerse,\ a dataset designed to enable systematic analysis of hallucinations across multiple languages, multiple generation tasks, and multiple hallucination categories. \HalluVerse\ covers four languages, English, Arabic, Hindi, and Turkish, and supports two generation tasks: question answering and dialogue summarization. The dataset explicitly distinguishes between entity-level, relation-level, and sentence-level hallucinations. Hallucinated outputs are constructed through a controlled editing process and validated by human annotators, ensuring clear alignment between original content and hallucinated generations. Using this dataset, we evaluate a diverse set of contemporary open-source and proprietary language models on fine-grained hallucination detection.
Our results show that question answering is consistently easier than dialogue summarization, while sentence-level hallucinations remain challenging even for the strongest models. Performance is highest in English and degrades in lower-resource languages, with Hindi exhibiting the lowest detection accuracy. Overall, \HalluVerse\ provides a realistic and challenging benchmark for studying hallucinations in multilingual, multi-task settings. We release the dataset 
to support future research on hallucination detection and mitigation\footnote{https://huggingface.co/datasets/sabdalja/HalluVerse-M3}.
\end{abstract}

\section{Introduction}

Large language models (LLMs) can generate coherent text across disciplines \citep{liu-etal-2024-learning,ramprasad-etal-2024-analyzing,zhu-etal-2024-multilingual}. However, they can also produce statements that lack factual grounding but are presented as plausible, a phenomenon widely known as hallucination. This problem poses significant challenges in mission-critical applications (healthcare, executive decision making, law) and may make specific user populations, especially those without domain expertise, more vulnerable to accepting fabricated claims as truth. The wide-scale adoption of LLMs is therefore, contingent on efficient hallucinations detection-an important problem in natural language processing research. Researchers have developed various datasets to analyze hallucinations in LLM generated text \citep{mubarak-etal-2024-halwasa,li-etal-2023-halueval,manakul-etal-2023-selfcheckgpt,mishra2024finegrained}. However, these datasets exhibit three primary limitations. First, they predominantly focus on English, which limits exploration of hallucination patterns in other languages. Second, most treat hallucination as a binary phenomenon that classify content as either hallucinated or factual, thereby obscuring the diversity of hallucination types observed in practice. Third, these resources typically address specific generation tasks, hindering systematic comparison of hallucinations across different application contexts.


To address these shortcomings, this work introduces \HalluVerse,\ a fine-grained, multilingual, and multi-task hallucination dataset designed to enable nuanced cross-linguistic and cross-task analysis of errors generated by LLMs. The main contributions of this work are as follows: (i) We introduce \HalluVerse,\ a multilingual dataset with fine-grained hallucination annotations spanning four languages: English, Arabic, Turkish, and Hindi. The dataset is designed to support systematic analysis of hallucination behavior across typologically and resource-diverse languages. (ii) We support two distinct natural language generation tasks, dialogue summarization and question answering, enabling controlled comparisons of hallucination characteristics across task settings with different levels of abstraction and grounding. (iii) We detail a comprehensive and reproducible dataset creation and annotation protocol. (iv) We perform a comprehensive analysis of state-of-the-art LLMs on our \HalluVerse\ dataset, establishing performance benchmarks across multiple languages and tasks. The dataset is publicly available to support continued research in this area. The dataset is publicly available to support continued research in this area. The remainder of the paper is organized as follows: Section \ref{sec:related} discusses relevant literature on hallucination datasets and benchmarks. Section \ref{sec:formal_background} provides theoretical background and notation. Section \ref{sec:method} details the dataset construction methodology. Section \ref{sec:analysis} presents a comprehensive analysis of the dataset. Section \ref{sec:exp} discusses experimental results of several LLMs on the dataset. Finally, Section \ref{sec:conclusion} concludes the paper and outlines future research directions.

\section{Related Works}

\label{sec:related}

\begin{table*}[htbp]
\centering
\caption{Comparison of HalluVerseM\textsuperscript{3} with existing hallucination datasets.}
\label{tab:dataset_comparison}
\begin{tabular}{lcccccccc}
\toprule
\textbf{Dataset} &
\textbf{Multilingual} &
\textbf{Multi-task} &
\textbf{Fine-grained} &
\textbf{Human Validation} \\
\midrule

HalluEval \citep{li-etal-2023-halueval} 
 & \ding{55} & \ding{51} & \ding{55} & \ding{55} \\

Wikibio \citep{manakul-etal-2023-selfcheckgpt} 
 & \ding{55} & \ding{55} & \ding{55} & \ding{51} \\

Halwasa \citep{mubarak-etal-2024-halwasa} 
 & \ding{55} & \ding{55} & \ding{55} & \ding{51} \\

FAVABENCH \citep{mishra2024finegrained} 
 & \ding{55} & \ding{55} & \ding{51} & \ding{55} \\

ANAH \citep{ji-etal-2024-anah}
 & \ding{51} & \ding{55} & \ding{55} & \ding{55} \\

HalluEntity \citep{yeh2025halluentity}
 & \ding{55} & \ding{55} & \ding{55} & \ding{51} \\

PsiloQA \citep{rykov-etal-2025-models}
 & \ding{51} & \ding{55} & \ding{55} & \ding{51} \\

\midrule
\textbf{HalluVerseM\textsuperscript{3}} (ours) 

& \ding{51} 
& \ding{51} 
& \ding{51}
& \ding{51} \\
\bottomrule
\end{tabular}
\end{table*}











\paragraph{Hallucination in Large Language Models.} The production of factually incorrect or unfaithful content by LLMs represents a significant barrier to their adoption in real-world applications \citep{li-etal-2024-dawn}. Academic interest in this issue has grown substantially in recent years \citep{10.1145/3703155,10.1145/3571730}. Research in this domain has progressed along two principal directions. The first examines the underlying causes and mechanisms responsible for hallucinated outputs \citep{mckenna-etal-2023-sources,das-etal-2022-diving}. The second develops methods for identifying and mitigating these errors \citep{chen2024inside,chen-etal-2024-unified-hallucination,varshney2023stitchtimesavesnine}. Detection approaches include the use of internal model representations with metrics such as Eigenscores \citep{chen2024inside} and the application of binary classification systems to distinguish factual from non-factual statements \citep{manakul-etal-2023-selfcheckgpt}. However, existing detection frameworks frequently operate at a coarse level of granularity or rely on proprietary language models, which constrains their generalizability and broader applicability \citep{wu2025survey}. Researchers have also studied unfaithful generations in specific domains such as text simplification \citep{devaraj2022evaluating} and summarization \citep{pagnoni2021understanding}. However, these categorizations are based on evaluating the faithfulness relative to a source document, and contrasts with LLM hallucinations which are based on world knowledge, where no explicit source text is available for verification.

\paragraph{Benchmarks and Datasets for Hallucination Evaluation.} Datasets have been created to facilitate the analysis of LLM-generated hallucinations. HaluEval is a collection of 35,000 question-answer pairs, summarization, and open-ended generation tasks in English \citep{li-etal-2023-halueval}. For Arabic, Halwasa provides 10,000 sentence-level annotations \citep{mubarak-etal-2024-halwasa}, and smaller resources such as the Wikibio hallucination dataset have also contributed to this area \citep{manakul-etal-2023-selfcheckgpt}. For fine-grained hallucinations, \citep{mishra2024finegrained} identifies three categories of factual errors, namely: entity-level errors, errors where an incorrect entity in a statement can be replaced to make it accurate; relational errors, resulting from inaccurate semantic links such as verbs or prepositions; and sentence-level errors, where large portions of a statement contradict available evidence. FAVABENCH is a dataset containing 1,000 annotated responses with fine-grained labels. ANAH offers a bilingual Chinese-English corpus with sentence-level annotations across a range of topics \citep{ji-etal-2024-anah}. More recently, datasets such as HalluEntity \citep{yeh2025halluentity} and PsiloQA \citep{rykov-etal-2025-models} have been introduced, extending the literature by targeting more specific forms of hallucinations and evaluation settings. Despite advancements, existing datasets are predominantly limited to English and/or single tasks, which restricts cross-linguistic analysis of hallucination patterns. On the other hand, multilingual datasets lack the fine-grained hallucination annotation. Moreover, most datasets do not address hallucination detection in multitask settings, for example, question answering or dialogue summarization, within a unified multilingual framework \citep{zhang2025poly}. Consequently, to bridge this gap, we present \HalluVerse,\ a fine-grained, multilingual, multitask dataset that provides hallucination annotations across four languages and three natural language generation tasks. Table~\ref{tab:dataset_comparison} presents a comparison between HalluVerseM\textsuperscript{3} and prior hallucination datasets, highlighting its unique combination of multilingual coverage, multi-task support, and fine-grained hallucination type annotations.

\section{Formal Background and Notation}
\label{sec:formal_background}

We formalize hallucinations in natural language generation to match the construction and evaluation protocol of \HalluVerse.\ The goal is to define (i) \emph{reference-consistency} of a model output and (ii) fine-grained hallucination types under controlled edits. Throughout, we treat $y^*$ as the benchmark grounding target; thus, ``hallucination'' in this work means content that is unsupported \emph{relative to the reference}.

\paragraph{Tasks, Languages, Inputs, and Reference Outputs.}
Let $\mathcal{L}=\{\textsc{En},\textsc{Ar},\textsc{Hi},\textsc{Tr}\}$ be the set of supported languages. We consider two generation tasks: question answering (\textsc{QA}) and dialogue summarization (\textsc{DS}). For each instance, the input is
\begin{equation}
x =
\begin{cases}
q, & \text{for \textsc{QA}},\\
d, & \text{for \textsc{DS}},
\end{cases}
\label{eq:input_def}
\end{equation}
where $q$ is a question and $d$ is a multi-turn dialogue. Each instance includes a human-verified \emph{reference output} $y^*$ (reference answer for \textsc{QA} or reference summary for \textsc{DS}), which serves as the grounding target.

\paragraph{Propositional Representation.}
Let $\mathcal{P}(y)$ denote the set of atomic propositions expressed in a text $y$. We represent a proposition as
\begin{equation}
p = \langle r;\, e_1,\ldots,e_k;\, a\rangle,
\label{eq:prop_tuple}
\end{equation}
where $r$ is a predicate (relation), $(e_1,\ldots,e_k)$ are entity arguments (named entities), and $a$ denotes optional attributes (e.g., dates, quantities, polarity), consistent with our annotation guidelines. This representation is conceptual (no semantic parser is required); it serves to define hallucination types, while labels are obtained via controlled edits and human validation.

\paragraph{Reference-Consistency and Hallucination.}
An output $y$ is \emph{reference-consistent} with respect to $y^*$ if
\begin{equation}
\mathcal{P}(y) \subseteq \mathcal{P}(y^*).
\label{eq:ref_consistency}
\end{equation}
An output contains a hallucination (relative to $y^*$) if
\begin{equation}
\exists\, p \in \mathcal{P}(y)\ \text{such that}\ p \notin \mathcal{P}(y^*).
\label{eq:hallucination}
\end{equation}

\paragraph{Controlled Hallucination Construction.}
We construct an edited output $\tilde{y}$ from $y^*$ by removing/altering some reference propositions and introducing unsupported propositions:
\begin{equation}
\mathcal{P}(\tilde{y})
=
\big(\mathcal{P}(y^*) \setminus \Delta^{-}\big)\ \cup\ \Delta^{+},
\label{eq:construction}
\end{equation}
where $\Delta^{-}$ denotes removed/altered propositions and $\Delta^{+}$ denotes newly introduced propositions. We enforce a single injected hallucination:
\begin{equation}
|\Delta^{+}| = 1,
\label{eq:single_injection}
\end{equation}
and require it to be unsupported by the reference:
\begin{equation}
\Delta^{+} \cap \mathcal{P}(y^*) = \emptyset.
\label{eq:unsupported}
\end{equation}
Let $\Delta^{+}=\{p^{+}\}$. We additionally enforce that edits in $\Delta^{-}$ do not introduce new unsupported propositions (deletions/paraphrases only), and we filter any instance that violates this constraint.

\paragraph{Alignment to Reference Propositions.}
Let $\operatorname{Align}(p, p')$ be a binary relation indicating that propositions $p$ and $p'$ refer to the same underlying fact slot/event up to a localized edit. For the injected proposition $p^{+}$, define its aligned reference set as
\begin{equation}
\mathcal{A}(p^{+}) = \{p' \in \mathcal{P}(y^*) : \operatorname{Align}(p^{+},p')\}.
\label{eq:aligned_set}
\end{equation}
In \HalluVerse,\ $\operatorname{Align}$ is operationalized by the controlled edit: $p^{+}$ is either minimally modified from a specific reference proposition (yielding $\mathcal{A}(p^{+})\neq\emptyset$) or introduced as a new proposition (yielding $\mathcal{A}(p^{+})=\emptyset$).

\paragraph{Fine-Grained Hallucination Types.}
Let $\mathcal{H}=\{\textsc{Entity},\textsc{Relation},\textsc{Sentence}\}$. Each instance has exactly one injected proposition $p^{+}$ and is assigned exactly one type. Under the operationalized alignment above and the enforced edit constraints, the cases below are mutually exclusive and collectively exhaustive.\\
\noindent
\textbf{Entity-level (\textsc{Entity}).} $p^{+}$ is \textsc{Entity} if there exists $p' \in \mathcal{A}(p^{+})$ such that predicate and attributes match, but at least one entity differs:
\begin{equation}
p^{+}=\langle r;\, e_1,\ldots,e_k;\, a\rangle,
\quad
p'=\langle r;\, e'_1,\ldots,e'_k;\, a\rangle,
\label{eq:entity_pair}
\end{equation}
and
\begin{equation}
\exists i \in \{1,\ldots,k\}\ \text{such that}\ e_i \neq e'_i.
\label{eq:entity_change}
\end{equation}

\paragraph{Relation-level (\textsc{Relation}).}
$p^{+}$ is \textsc{Relation} if there exists $p' \in \mathcal{A}(p^{+})$ such that entities match, but predicate and/or attributes differ:
\begin{equation}
p^{+}=\langle r;\, e_1,\ldots,e_k;\, a\rangle,
\quad
p'=\langle r';\, e_1,\ldots,e_k;\, a'\rangle,
\label{eq:relation_pair}
\end{equation}
with
\begin{equation}
(r,a)\neq (r',a').
\label{eq:relation_change}
\end{equation}

\paragraph{Sentence-level (\textsc{Sentence}).}
$p^{+}$ is \textsc{Sentence} if it has no aligned reference proposition, i.e., it is an \emph{unaligned/unanchored statement} with respect to the reference:
\begin{equation}
\mathcal{A}(p^{+}) = \emptyset.
\label{eq:sentence_def}
\end{equation}

\paragraph{Hallucination Detection as Structured Prediction.}

Given $(y^*, \tilde{y})$, define the introduced semantic difference
\begin{equation}
\Delta_{\mathrm{diff}} = \mathcal{P}(\tilde{y}) \setminus \mathcal{P}(y^*).
\label{eq:diff}
\end{equation}
Under the constraints above, the newly introduced unsupported content is unique and
\begin{equation}
\Delta_{\mathrm{diff}} = \{p^{+}\}.
\label{eq:diff_singleton}
\end{equation}
A detector $f_{\theta}$ predicts the hallucination type:
\begin{equation}
f_{\theta}(y^*, \tilde{y}) \mapsto \hat{h},
\quad \hat{h} \in \mathcal{H}.
\label{eq:classifier}
\end{equation}

\section{Constructing \HalluVerse}
\label{sec:method}

The goal of this dataset is to support a systematic and multilingual study of hallucinations in natural language generation across different tasks. The dataset is designed to capture how hallucinations manifest in several contexts, covering question answering and dialogue summarization. Each instance in the dataset consists of an original ground-truth output, a hallucinated version of that output, and an explicit label identifying the hallucination type. The construction process follows a structured pipeline that includes task-specific data selection, multilingual data preparation, automatic hallucination injection, and human validation.

\subsection{Task and Data Selection}

\paragraph{Question Answering.}
For the question answering task, we rely on an English factual question answering dataset introduced by \citet{lin2022truthfulqameasuringmodelsmimic}, which was originally constructed to evaluate factual consistency in long-form answers. The dataset contains questions paired with reference answers grounded in verifiable evidence. We use the reference answers as the factual basis for hallucination injection. Only instances with clearly stated, self-contained answers were retained, as these are most suitable for controlled hallucination generation and annotation.

\paragraph{Dialogue Summarization.}
For the summarization task, we use the DialogSum dataset \citep{chen-etal-2021-dialogsum}, a widely used benchmark for dialogue summarization that consists of multi-turn conversations paired with human-written summaries. We focus on summaries that describe concrete events, entities, or outcomes discussed in the dialogue, since these provide a reliable foundation for identifying and injecting hallucinated content. 

\subsection{Multilingual Data Preparation}

The dataset covers four languages: English, Arabic, Hindi, and Turkish. English serves as the source language for both tasks. For the remaining languages, we create parallel task outputs through automatic translation followed by manual verification. Translation is performed using the Google Translate API, applied at the sentence level for question answering outputs and at the summary level for dialogue summarization outputs. This granularity helps preserve semantic content and structure across tasks. Following translation, all instances are reviewed by native speakers to verify grammatical correctness, naturalness, and semantic faithfulness to the source text. Instances that exhibit mistranslations, unnatural phrasing, or shifts in meaning  are excluded. This process results in high-quality multilingual data that supports a reliable cross-lingual analysis of hallucinations.

\subsection{Automated Hallucination Injection}

To generate hallucinated outputs, we adopt a controlled automatic editing process that introduces fine-grained hallucinations while preserving overall fluency. Each original output is paired with a hallucinated version created by modifying a specific aspect of the content. We consider three hallucination types: entity hallucinations, relation hallucinations, and sentence-level hallucinations. Entity hallucinations involve substituting a named entity with another plausible but incorrect entity. Relation hallucinations alter factual relationships, such as temporal order, causality, or attribution. Sentence-level hallucinations introduce information that is unsupported or entirely fabricated within the context. For each instance, the hallucination type is selected using a randomized assignment procedure to ensure balanced coverage across types. Hallucination injection is performed using an LLM prompted with a structured instruction format. The prompt explicitly specifies the hallucination type to be injected, provides a concise definition of that type, and includes illustrative examples of original and hallucinated outputs. This design ensures that the generated hallucinations adhere closely to the intended category while maintaining grammatical and stylistic consistency with the original text.



\subsection{Human Annotation}

Human annotation is used to validate both the presence and the type of hallucination in the edited outputs. For each language, two native speakers with at least a bachelor’s degree independently annotate the data. Annotators are presented with pairs consisting of the original output and its edited version, without access to the hallucination label or any additional metadata. Before annotation, annotators participate in a structured training session and are provided with a detailed annotation protocol that defines each hallucination type and includes task-specific examples. Annotators assign a hallucination label if the edited output contains a clear hallucination and assign a null label if the output is grammatically incorrect or does not introduce any hallucinated content. Inter-annotator agreement is measured using Cohen’s Kappa \cite{kappa}. We observe substantial to near-perfect agreement across all languages, indicating consistent understanding of the annotation guidelines. Specifically, the average agreement is $\kappa=0.83$ for English, $\kappa=0.79$ for Arabic, $\kappa=0.76$ for Turkish, and $\kappa=0.74$ for Hindi, averaged across both question answering and dialogue summarization tasks. These results reflect both the inherent difficulty of fine-grained hallucination annotation and differences in linguistic complexity across languages. Disagreements are resolved through discussion, with annotators jointly reviewing each instance and reaching consensus on a final label. The resulting annotations form the final validated version of the dataset.

\section{Data Analysis}
\label{sec:analysis}
This section presents an analysis of the constructed dataset across tasks, languages, and hallucination types. The analysis focuses on task coverage, multilingual balance, hallucination type distributions, and basic textual statistics. Together, these analyses provide insights into the diversity and structural properties of the dataset, and help characterize its suitability for studying hallucinations in multilingual and multi-task settings.

\begin{figure}[t]
\centering
\caption{
Dataset distribution after filtering \kval{0}-labeled instances.
Stacked bars show QA and Dialogue Summarization (Summ.) counts per language; segment counts and totals are annotated.
}
\label{fig:data_dist_stacked_singlecol_framed}

\resizebox{0.99\columnwidth}{!}{%
\begin{tikzpicture}[x=0.0093cm,y=0.78cm, font=\small]

  \tikzset{
    qa/.style={fill=blue!56, draw=black!25, line width=0.25pt},
    su/.style={fill=blue!26, draw=black!25, line width=0.25pt},
    lab/.style={anchor=east, font=\small},
    segnum/.style={font=\scriptsize, text=black!95},
    total/.style={anchor=west, font=\scriptsize},
    legendtxt/.style={anchor=west, font=\scriptsize},
    meta/.style={anchor=west, font=\scriptsize, text=black!85}
  }

  \draw[draw=black!35, rounded corners=2pt, line width=0.35pt]
    (-170,-0.08) rectangle (1265,6.3);

  \draw[draw=black!20, rounded corners=2pt, fill=black!2, line width=0.25pt]
    (-80,4.3) rectangle (1225,6);

  \node[meta] at (-60,5.5) {%
    \textbf{Total:} 4038 \hspace{12pt}
    \textbf{Languages:} 4 \hspace{12pt}
    \textbf{Tasks:} 2
  };

  \node[meta] at (-60,4.7) {
    \textbf{QA:} 2885 (71.4\%) \hspace{14pt}
    \textbf{Summ.:} 1153 (28.6\%)
  };

  \begin{scope}[shift={(800,4.95)}]
    \draw[qa] (0,0) rectangle +(90,0.27);
    \node[legendtxt] at (100,0.135) {QA};

    \draw[su] (175,0) rectangle +(90,0.27);
    \node[legendtxt] at (275,0.135) {Summ.};
  \end{scope}


  \node[lab] at (-25,3.70) {English};
  \draw[qa] (0,3.45) rectangle +(779,0.55);
  \draw[su] (779,3.45) rectangle +(297,0.55);
  \node[segnum] at (389.5,3.73) {779};
  \node[segnum] at (927.5,3.73) {297};
  \node[total]  at (1100,3.73) {\textbf{1076}};

  \node[lab] at (-25,2.70) {Arabic};
  \draw[qa] (0,2.45) rectangle +(780,0.55);
  \draw[su] (780,2.45) rectangle +(300,0.55);
  \node[segnum] at (390,2.73) {780};
  \node[segnum] at (930,2.73) {300};
  \node[total]  at (1100,2.73) {\textbf{1080}};

  \node[lab] at (-25,1.70) {Hindi};
  \draw[qa] (0,1.45) rectangle +(575,0.55);
  \draw[su] (575,1.45) rectangle +(261,0.55);
  \node[segnum] at (287.5,1.73) {575};
  \node[segnum] at (705.5,1.73) {261};
  \node[total]  at (860,1.73) {\textbf{836}};

  \node[lab] at (-25,0.70) {Turkish};
  \draw[qa] (0,0.45) rectangle +(751,0.55);
  \draw[su] (751,0.45) rectangle +(295,0.55);
  \node[segnum] at (375.5,0.73) {751};
  \node[segnum] at (898.5,0.73) {295};
  \node[total]  at (1070,0.73) {\textbf{1046}};

\end{tikzpicture}%
}
\end{figure}
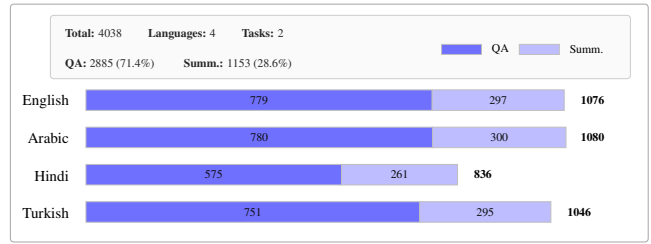


\subsection{Task and Language Coverage}

The dataset consists of two generation tasks, namely question answering and dialogue summarization, and covers four languages: English, Arabic, Hindi, and Turkish. Figure \ref{fig:data_dist_stacked_singlecol_framed} illustrates the proportion of instances for each task-language combination, after filtering out the data instances labeled as `0' by the annotators. As shown in Figure~\ref{fig:data_dist_stacked_singlecol_framed}, both tasks are represented across all four languages, enabling consistent analysis of hallucination behavior across tasks and linguistic contexts. English constitutes the largest portion of the dataset, reflecting its role as the source language, while the remaining languages are more evenly distributed. The number of question answering instances in Hindi is lower than in the other languages, primarily due to grammatical errors introduced during hallucination injection, which led to the exclusion of affected instances according to the annotation protocol. This distribution supports both within-language and cross-lingual analyses of hallucination patterns while maintaining annotation quality.

\begin{figure*}[htbp]
\centering
\begin{subfigure}[b]{0.24\textwidth}
\centering
\includegraphics[width=\textwidth]{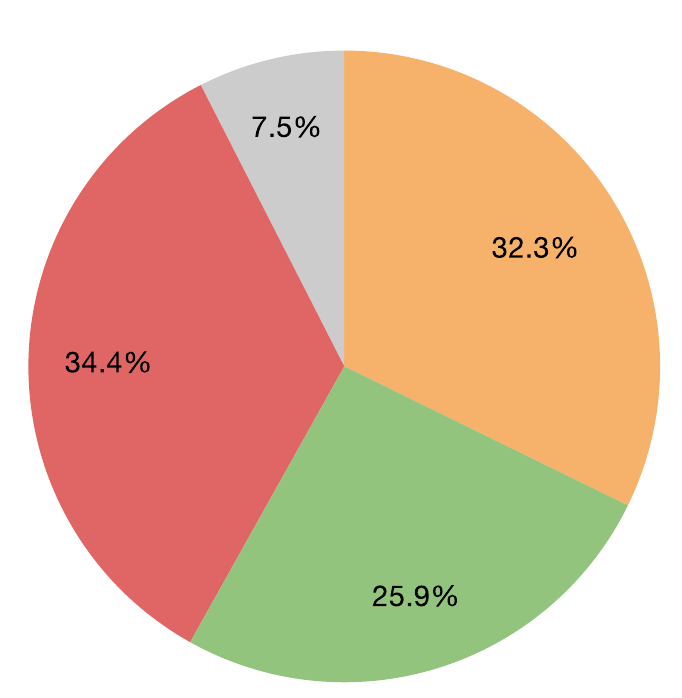}
\caption{English}
\label{fig:en_dist}
\end{subfigure}
\hfill
\begin{subfigure}[b]{0.24\textwidth}
\centering
\includegraphics[width=\textwidth]{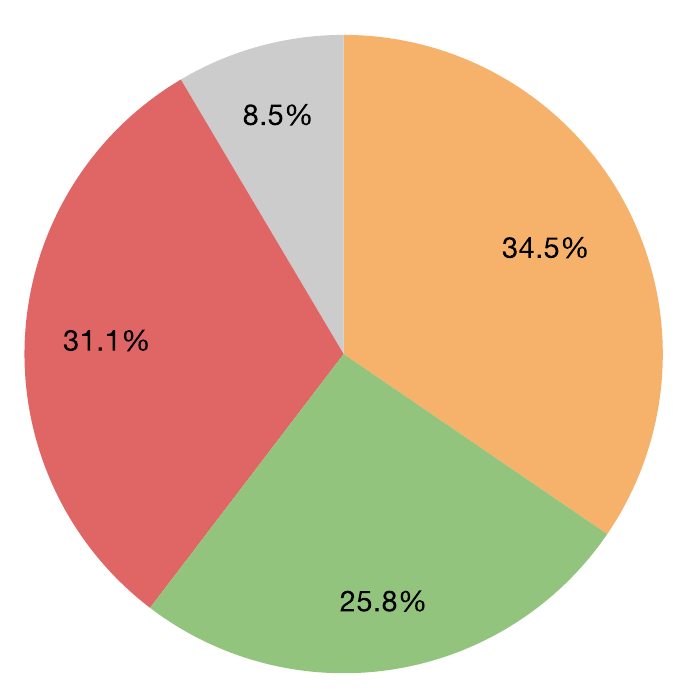}
\caption{Arabic}
\label{fig:ar_dist}
\end{subfigure}
\hfill
\begin{subfigure}[b]{0.24\textwidth}
\centering
\includegraphics[width=\textwidth]{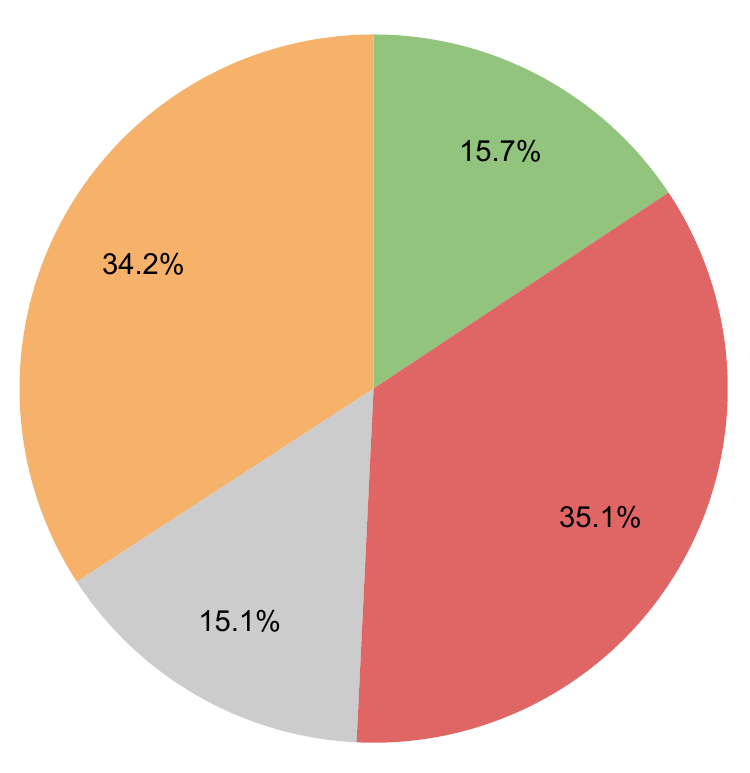}
\caption{Hindi}
\label{fig:hi_dist}
\end{subfigure}
\hfill
\begin{subfigure}[b]{0.24\textwidth}
\centering
\includegraphics[width=\textwidth]{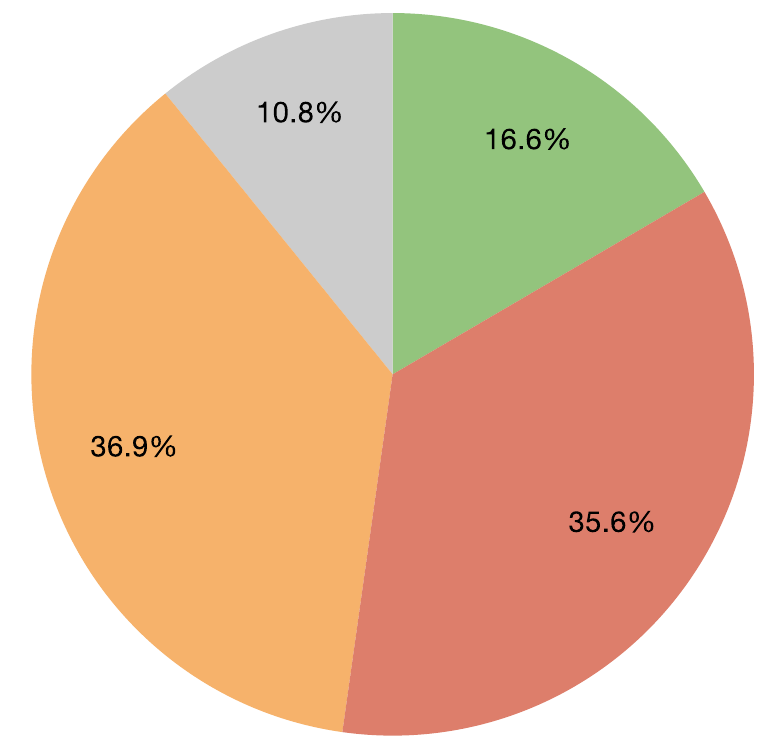}
\caption{Turkish}
\label{fig:tr_dist}
\end{subfigure}
\caption{Hallucination type distribution across languages aggregated across both tasks. Entity (red), relation (green), sentence (orange), \kval{0} (grey).}
\label{fig:hallucination_dist}
\end{figure*}

\subsection{Hallucination Type Distribution}

Each hallucinated instance is annotated as one of three types: entity hallucination, relation hallucination, or sentence-level hallucination. Instances labeled as `\kval{0}' are also represented in the analysis.  Fig. \ref{fig:hallucination_dist} presents the hallucination type distribution for each language, aggregated across tasks. 
Overall, the distributions show a good balance among hallucinations at the entity, relation, and sentence-level for English and Arabic, with no single category dominating the data. In contrast, for Turkish and Hindi, a higher proportion of instances is labeled as \kval{0}. This difference is largely due to grammatical errors introduced during the construction of relation-level hallucinations. As specified in the annotation protocol, edited outputs that contained grammatical inconsistencies or unnatural constructions were labeled as \kval{0} and excluded from the hallucination categories. Relation-level edits were particularly prone to such errors in Turkish and Hindi, leading to a lower observed proportion of relation hallucinations in these languages. Despite this effect, entity- and sentence-level hallucinations remain well represented across all languages, ensuring sufficient coverage for fine-grained analysis while reflecting realistic linguistic constraints encountered during dataset creation.

\subsection{Discussion}

Taken together, these analyses demonstrate that the dataset is balanced across tasks, languages, and hallucination types. The presence of consistent patterns across languages, alongside meaningful task-specific differences, highlights the dataset’s suitability for studying hallucination behavior in multilingual and multi-task generation settings. The diversity observed in both structure and content supports a wide range of experimental setups, including hallucination detection, classification, and cross-lingual transfer.

\section{Experiments}
\label{sec:exp}

\HalluVerse\ enables systematic evaluation of hallucination detection across tasks, languages, and model families. In this section, we benchmark a diverse set of LLMs on fine-grained hallucination detection, focusing on multilingual question answering and dialogue summarization. All experiments are conducted using prompting only, without any model fine-tuning or access to internal representations.

\subsection{Task Setup}

Given a pair consisting of an original task output and a hallucinated version, models are instructed to identify the type of hallucination present. Predictions are made among three categories: entity-level, relation-level, and sentence-level hallucinations. Evaluation is performed independently for question answering and summarization, and accuracy is reported as the primary metric. 
\subsection{Evaluation Models}

We evaluate a broad range of contemporary models, including both open-source and proprietary systems. Open-source models include Phi-4 (14B), Mistral-7B-Instruct, Gemma-2-27B, Qwen-2.5 (7B, 32B, and 72B variants), LLaMA-3.3-70B, and DeepSeek-V2.5. Proprietary models include PaLM 2, Claude-3.5 Sonnet, Gemini-2.5-Pro, GPT-4o Mini, GPT-4.1, and GPT-4o. This selection reflects a range of model scales, training paradigms, and multilingual capabilities, allowing for a comprehensive evaluation. All models are evaluated using the same prompt template and the same number of in-context examples per hallucination type, ensuring a fair comparison.

\subsection{Results}

\begin{table}[t]
\centering
\caption{
Hallucination-detection accuracy (\%) across tasks (QA, Summ.) and languages (En/Ar/Hi/Tr), averaged over hallucination types.
Cell shading indicates performance (darker is better); best results per column are \textbf{bold}.
}
\label{tab:mt_results_heatmap_singlecol}

\scriptsize
\setlength{\tabcolsep}{3.2pt}
\renewcommand{\arraystretch}{1.10}

\newcommand{\m}[1]{#1}

\begin{tabular}{p{2.35cm}cccc@{\hspace{3pt}}cccc}
\toprule
\textbf{Model} &
\multicolumn{4}{c}{\textbf{QA}} &
\multicolumn{4}{c}{\textbf{Summ.}} \\
\cmidrule(lr){2-5}\cmidrule(lr){6-9}
& \textbf{En} & \textbf{Ar} & \textbf{Hi} & \textbf{Tr}
& \textbf{En} & \textbf{Ar} & \textbf{Hi} & \textbf{Tr} \\
\midrule
\m{Phi-4 (14B)}           & \hmE{77.9} & \hmD{74.0} & \hmD{70.9} & \hmD{73.8} & \hmD{70.8} & \hmC{66.8} & \hmC{62.6} & \hmC{66.2} \\
\m{Mistral-7B}            & \hmB{65.5} & \hmB{61.5} & \hmB{56.7} & \hmB{60.3} & \hmB{56.1} & \hmA{51.4} & \hmA{46.8} & \hmA{50.2} \\
\m{Qwen-2.5-7B}           & \hmB{68.5} & \hmB{64.8} & \hmB{60.2} & \hmB{63.7} & \hmA{59.4} & \hmA{54.9} & \hmA{50.1} & \hmA{53.6} \\
\m{Gemma-2 (27B)}         & \hmD{71.2} & \hmC{67.3} & \hmC{62.9} & \hmC{66.5} & \hmC{62.1} & \hmB{57.6} & \hmA{52.8} & \hmB{56.4} \\
\m{Qwen-2.5-32B}          & \hmE{75.9} & \hmD{72.1} & \hmC{67.8} & \hmD{71.3} & \hmC{66.8} & \hmC{62.4} & \hmB{57.4} & \hmB{61.0} \\
\m{Qwen-2.5-72B}          & \hmE{78.8} & \hmE{75.1} & \hmD{71.0} & \hmD{74.4} & \hmD{69.7} & \hmC{65.3} & \hmB{60.3} & \hmC{63.9} \\
\m{LLaMA-3.3 (70B)}       & \hmE{79.5} & \hmE{75.8} & \hmD{71.6} & \hmE{75.1} & \hmD{70.3} & \hmC{66.0} & \hmC{61.2} & \hmC{64.7} \\
\m{DeepSeek-V2.5}         & \hmF{82.8} & \hmE{79.1} & \hmE{75.0} & \hmE{78.4} & \hmD{73.5} & \hmD{69.3} & \hmC{64.2} & \hmC{67.8} \\
\midrule
\m{PaLM 2}                & \hmE{78.9} & \hmE{75.2} & \hmD{71.1} & \hmD{74.6} & \hmD{69.8} & \hmC{65.4} & \hmB{60.6} & \hmC{64.2} \\
\m{Claude-3.5}            & \hmF{85.9} & \hmF{82.1} & \hmE{78.0} & \hmF{81.5} & \hmE{76.8} & \hmD{72.5} & \hmC{67.8} & \hmD{71.2} \\
\m{Gemini-2.5}            & \hmF{85.2} & \hmF{81.4} & \hmE{77.2} & \hmE{80.7} & \hmE{76.1} & \hmD{71.8} & \hmC{67.1} & \hmD{70.6} \\
\m{GPT-4o mini}           & \hmF{82.3} & \hmE{78.6} & \hmD{74.4} & \hmE{77.9} & \hmD{73.2} & \hmD{69.0} & \hmC{64.1} & \hmC{67.6} \\
\m{GPT-4.1}               & \hmF{87.1} & \hmF{83.6} & \hmF{79.7} & \hmF{83.0} & \hmE{78.6} & \hmD{74.3} & \hmD{69.8} & \hmD{73.2} \\
\m{GPT-4o}                & \best{\hmF{88.6}} & \best{\hmF{85.1}} & \best{\hmF{81.2}} & \best{\hmF{84.6}}
                          & \best{\hmE{80.2}} & \best{\hmE{76.0}} & \best{\hmD{71.6}} & \best{\hmE{75.1}} \\
\bottomrule
\end{tabular}

\vspace{0.8mm}
\centering
\tiny
\setlength{\tabcolsep}{2.6pt}
\renewcommand{\arraystretch}{1.0}
\begin{tabular}{lcccccc}
 &
\legbox{blue!6}{\ensuremath{<55}} &
\legbox{blue!14}{\ensuremath{55\mbox{--}<62}} &
\legbox{blue!26}{\ensuremath{62\mbox{--}<69}} &
\legbox{blue!40}{\ensuremath{69\mbox{--}<75}} &
\legbox{blue!56}{\ensuremath{75\mbox{--}<81}} &
\legbox{blue!72}{\ensuremath{\ge 81}} \\
\end{tabular}

\end{table}

\paragraph{Overall Performance.}
Table \ref{tab:mt_results_heatmap_singlecol} summarizes hallucination detection accuracy across tasks and languages. Across all models, question answering consistently yields higher accuracy than dialogue summarization. This trend reflects the more localized nature of hallucinations in question answering, where factual errors often involve identifiable entities or relations, as opposed to summarization, where hallucinations are more diffuse and embedded within longer spans of text. Among all evaluated models, GPT-4o achieves the strongest overall performance, with accuracy reaching close to \kval{89\%} on question answering and exceeding \kval{80\%} on summarization in English, while maintaining similarly strong performance across the remaining languages. This outcome is expected, as GPT-4o is also used in the controlled generation process to inject hallucinations, which may confer an advantage in recognizing the corresponding error patterns. GPT-4.1 and Claude-3.5 Sonnet closely follow, indicating that recent frontier models exhibit improved robustness to hallucinated content across both tasks and languages.

\begin{table}[t]
\centering
\caption{
English (EN): hallucination-detection accuracy (\%) by hallucination type for QA and summarization.
Cell shading indicates performance (darker is better); best per column is \textbf{bold}.
}
\label{tab:hallucination_en_heatmap}

\scriptsize
\setlength{\tabcolsep}{4.2pt}
\renewcommand{\arraystretch}{1.12}

\begin{tabular}{lccc@{\hspace{4pt}}ccc}
\toprule
\textbf{Model} &
\multicolumn{3}{c}{\textbf{QA}} &
\multicolumn{3}{c}{\textbf{Summ.}} \\
\cmidrule(lr){2-4}\cmidrule(lr){5-7}
& \textbf{Ent.} & \textbf{Rel.} & \textbf{Sent.} & \textbf{Ent.} & \textbf{Rel.} & \textbf{Sent.} \\
\midrule
Phi-4 (14B)              & \hmF{90.2} & \hmD{78.6} & \hmC{65.1} & \hmE{84.1} & \hmD{70.4} & \hmB{57.8} \\
Mistral-7B Instruct      & \hmC{77.5} & \hmC{66.2} & \hmB{52.9} & \hmC{69.8} & \hmB{58.4} & \hmA{45.1} \\
Qwen-2.5-7B              & \hmE{80.6} & \hmC{69.1} & \hmB{55.9} & \hmD{72.3} & \hmC{60.2} & \hmA{47.2} \\
Gemma-2 27B              & \hmE{82.9} & \hmD{71.4} & \hmB{58.0} & \hmD{74.8} & \hmC{62.6} & \hmA{49.6} \\
Qwen-2.5-32B             & \hmE{86.8} & \hmD{75.2} & \hmC{61.7} & \hmE{79.2} & \hmC{67.1} & \hmB{54.9} \\
Qwen-2.5-72B             & \hmE{89.1} & \hmD{77.6} & \hmC{63.8} & \hmE{81.4} & \hmC{69.4} & \hmB{56.8} \\
LLaMA-3.3-70B            & \hmE{89.8} & \hmD{78.0} & \hmC{63.6} & \hmE{82.2} & \hmD{70.1} & \hmB{56.3} \\
DeepSeek-V2.5            & \hmF{91.6} & \hmE{80.4} & \hmC{66.3} & \hmE{84.0} & \hmD{72.2} & \hmB{58.6} \\
\midrule
PaLM 2                   & \hmE{88.9} & \hmD{77.1} & \hmC{62.7} & \hmE{81.0} & \hmC{69.0} & \hmB{55.2} \\
Claude-3.5 Sonnet        & \hmF{94.1} & \hmE{83.5} & \hmD{70.2} & \hmE{86.9} & \hmD{75.1} & \hmC{63.4} \\
Gemini-2.5-Pro           & \hmF{93.6} & \hmE{82.8} & \hmC{69.4} & \hmE{86.2} & \hmD{74.4} & \hmC{62.6} \\
GPT-4o Mini              & \hmF{91.3} & \hmD{79.9} & \hmC{66.7} & \hmE{84.7} & \hmD{72.0} & \hmB{59.2} \\
GPT-4.1                  & \hmF{95.0} & \hmE{85.1} & \hmD{72.8} & \hmE{88.6} & \hmD{77.3} & \hmC{65.9} \\
GPT-4o                   & \best{\hmF{96.1}} & \best{\hmE{86.7}} & \best{\hmD{74.4}} & \best{\hmF{90.1}} & \best{\hmD{79.0}} & \best{\hmC{67.6}} \\
\bottomrule
\end{tabular}

\vspace{1.1mm}
\centering
\scriptsize
\setlength{\tabcolsep}{3.2pt}
\renewcommand{\arraystretch}{1.05}
\begin{tabular}{lcccccc}
 &
\legbox{blue!6}{\ensuremath{<50}} &
\legbox{blue!14}{\ensuremath{50\mbox{--}<60}} &
\legbox{blue!26}{\ensuremath{60\mbox{--}<70}} &
\legbox{blue!40}{\ensuremath{70\mbox{--}<80}} &
\legbox{blue!56}{\ensuremath{80\mbox{--}<90}} &
\legbox{blue!72}{\ensuremath{\ge 90}} \\
\end{tabular}

\end{table}

\begin{table}[t]
\centering
\caption{
Arabic (AR): hallucination-detection accuracy (\%) by hallucination type for QA and summarization.
Cell shading indicates performance (darker is better); best results per column are \textbf{bold}.
}
\label{tab:hallucination_ar_heatmap}

\scriptsize
\setlength{\tabcolsep}{4.2pt}
\renewcommand{\arraystretch}{1.12}

\begin{tabular}{lccc@{\hspace{4pt}}ccc}
\toprule
\textbf{Model} &
\multicolumn{3}{c}{\textbf{QA}} &
\multicolumn{3}{c}{\textbf{Summ.}} \\
\cmidrule(lr){2-4}\cmidrule(lr){5-7}
& \textbf{Ent.} & \textbf{Rel.} & \textbf{Sent.}
& \textbf{Ent.} & \textbf{Rel.} & \textbf{Sent.} \\
\midrule
Phi-4 (14B)              & \hmE{87.0} & \hmD{75.1} & \hmC{62.4} & \hmE{80.1} & \hmC{67.3} & \hmB{54.6} \\
Mistral-7B Instruct      & \hmD{73.1} & \hmC{62.4} & \hmA{48.9} & \hmC{65.2} & \hmB{54.3} & \hmA{41.7} \\
Qwen-2.5-7B              & \hmD{76.4} & \hmC{65.8} & \hmB{51.6} & \hmC{68.1} & \hmB{57.2} & \hmA{43.4} \\
Gemma-2 27B              & \hmD{79.0} & \hmC{68.3} & \hmB{54.4} & \hmD{70.9} & \hmB{59.6} & \hmA{46.1} \\
Qwen-2.5-32B             & \hmE{83.5} & \hmD{72.4} & \hmB{59.6} & \hmD{75.4} & \hmC{64.2} & \hmB{51.8} \\
Qwen-2.5-72B             & \hmE{85.7} & \hmD{74.9} & \hmC{61.7} & \hmD{77.8} & \hmC{66.8} & \hmB{53.9} \\
LLaMA-3.3-70B            & \hmE{86.4} & \hmD{74.9} & \hmC{61.8} & \hmD{78.7} & \hmC{67.0} & \hmB{54.1} \\
DeepSeek-V2.5            & \hmE{88.2} & \hmD{77.0} & \hmC{64.1} & \hmE{80.5} & \hmC{69.1} & \hmB{56.4} \\
\midrule
PaLM 2                   & \hmE{85.2} & \hmD{74.0} & \hmC{60.6} & \hmD{77.5} & \hmC{66.1} & \hmB{53.0} \\
Claude-3.5 Sonnet        & \hmF{90.7} & \hmD{79.6} & \hmC{67.3} & \hmE{83.4} & \hmD{71.8} & \hmB{59.7} \\
Gemini-2.5-Pro           & \hmF{90.1} & \hmD{78.9} & \hmC{66.5} & \hmE{82.6} & \hmD{71.1} & \hmB{58.9} \\
GPT-4o Mini              & \hmE{88.0} & \hmD{77.4} & \hmC{64.9} & \hmE{81.1} & \hmC{69.8} & \hmB{57.2} \\
GPT-4.1                  & \hmF{92.0} & \hmE{81.4} & \hmD{69.8} & \hmE{85.6} & \hmD{74.0} & \hmC{62.5} \\
GPT-4o                   & \best{\hmF{93.4}} & \best{\hmE{82.9}} & \best{\hmD{71.5}}
                         & \best{\hmE{87.2}} & \best{\hmD{75.6}} & \best{\hmC{64.1}} \\
\bottomrule
\end{tabular}

\vspace{1.0mm}
\centering
\scriptsize
\setlength{\tabcolsep}{3.2pt}
\renewcommand{\arraystretch}{1.05}
\begin{tabular}{lcccccc}
 &
\legbox{blue!6}{\ensuremath{<50}} &
\legbox{blue!14}{\ensuremath{50\mbox{--}<60}} &
\legbox{blue!26}{\ensuremath{60\mbox{--}<70}} &
\legbox{blue!40}{\ensuremath{70\mbox{--}<80}} &
\legbox{blue!56}{\ensuremath{80\mbox{--}<90}} &
\legbox{blue!72}{\ensuremath{\ge 90}} \\
\end{tabular}
\end{table}

\begin{table}[t]
\centering
\caption{
Hindi (HI): hallucination-detection accuracy (\%) by hallucination type for QA and summarization.
Cell shading indicates performance (darker is better); best results per column are \textbf{bold}.
}
\label{tab:hallucination_hi_heatmap}

\scriptsize
\setlength{\tabcolsep}{4.2pt}
\renewcommand{\arraystretch}{1.12}

\begin{tabular}{lccc@{\hspace{4pt}}ccc}
\toprule
\textbf{Model} &
\multicolumn{3}{c}{\textbf{QA}} &
\multicolumn{3}{c}{\textbf{Summ.}} \\
\cmidrule(lr){2-4}\cmidrule(lr){5-7}
& \textbf{Ent.} & \textbf{Rel.} & \textbf{Sent.}
& \textbf{Ent.} & \textbf{Rel.} & \textbf{Sent.} \\
\midrule
Phi-4 (14B)              & \hmE{83.1} & \hmD{71.4} & \hmB{58.2} & \hmD{75.0} & \hmC{63.1} & \hmA{49.8} \\
Mistral-7B Instruct      & \hmC{68.9} & \hmB{57.6} & \hmA{44.8} & \hmB{59.4} & \hmA{48.2} & \hmA{35.6} \\
Qwen-2.5-7B              & \hmD{72.1} & \hmC{60.9} & \hmA{47.5} & \hmC{62.0} & \hmB{50.4} & \hmA{37.8} \\
Gemma-2 27B              & \hmD{74.2} & \hmC{63.1} & \hmA{49.6} & \hmC{64.3} & \hmB{52.7} & \hmA{39.9} \\
Qwen-2.5-32B             & \hmD{78.1} & \hmC{66.8} & \hmB{54.3} & \hmC{69.2} & \hmB{58.0} & \hmA{45.1} \\
Qwen-2.5-72B             & \hmE{80.6} & \hmC{69.4} & \hmB{56.8} & \hmD{71.8} & \hmC{60.6} & \hmA{47.6} \\
LLaMA-3.3-70B            & \hmE{81.0} & \hmC{69.5} & \hmB{56.9} & \hmD{72.2} & \hmC{60.3} & \hmA{47.6} \\
DeepSeek-V2.5            & \hmE{83.0} & \hmD{71.8} & \hmB{59.1} & \hmD{74.0} & \hmC{62.5} & \hmB{50.7} \\
\midrule
PaLM 2                   & \hmE{80.0} & \hmC{68.7} & \hmB{55.4} & \hmD{71.0} & \hmB{59.8} & \hmA{46.5} \\
Claude-3.5 Sonnet        & \hmE{86.2} & \hmD{75.3} & \hmC{63.0} & \hmD{77.8} & \hmC{66.4} & \hmB{54.9} \\
Gemini-2.5-Pro           & \hmE{85.5} & \hmD{74.6} & \hmC{62.2} & \hmD{77.0} & \hmC{65.7} & \hmB{54.1} \\
GPT-4o Mini              & \hmE{83.2} & \hmD{72.1} & \hmB{59.7} & \hmD{75.2} & \hmC{63.8} & \hmB{51.4} \\
GPT-4.1                  & \hmE{88.3} & \hmD{77.6} & \hmC{65.8} & \hmE{80.5} & \hmC{69.3} & \hmB{57.6} \\
GPT-4o                   & \best{\hmE{89.6}} & \best{\hmD{79.0}} & \best{\hmC{67.4}}
                         & \best{\hmE{82.0}} & \best{\hmD{71.1}} & \best{\hmB{59.8}} \\
\bottomrule
\end{tabular}

\vspace{1.0mm}
\centering
\scriptsize
\setlength{\tabcolsep}{3.2pt}
\renewcommand{\arraystretch}{1.05}
\begin{tabular}{lcccccc}
 &
\legbox{blue!6}{\ensuremath{<50}} &
\legbox{blue!14}{\ensuremath{50\mbox{--}<60}} &
\legbox{blue!26}{\ensuremath{60\mbox{--}<70}} &
\legbox{blue!40}{\ensuremath{70\mbox{--}<80}} &
\legbox{blue!56}{\ensuremath{80\mbox{--}<90}} &
\legbox{blue!72}{\ensuremath{\ge 90}} \\
\end{tabular}
\end{table}

\begin{table}[t]
\centering
\caption{
Turkish (TR): hallucination-detection accuracy (\%) by hallucination type for QA and summarization.
Cell shading indicates performance (darker is better); best results per column are \textbf{bold}.
}
\label{tab:hallucination_tr_heatmap}

\scriptsize
\setlength{\tabcolsep}{4.2pt}
\renewcommand{\arraystretch}{1.12}

\begin{tabular}{lccc@{\hspace{4pt}}ccc}
\toprule
\textbf{Model} &
\multicolumn{3}{c}{\textbf{QA}} &
\multicolumn{3}{c}{\textbf{Summ.}} \\
\cmidrule(lr){2-4}\cmidrule(lr){5-7}
& \textbf{Ent.} & \textbf{Rel.} & \textbf{Sent.}
& \textbf{Ent.} & \textbf{Rel.} & \textbf{Sent.} \\
\midrule
Phi-4 (14B)              & \hmE{86.1} & \hmD{74.6} & \hmC{60.8} & \hmD{78.9} & \hmC{66.8} & \hmB{52.9} \\
Mistral-7B Instruct      & \hmD{72.4} & \hmC{61.1} & \hmA{47.6} & \hmC{64.5} & \hmB{53.4} & \hmA{40.9} \\
Qwen-2.5-7B              & \hmD{75.7} & \hmC{64.3} & \hmB{50.8} & \hmC{67.6} & \hmB{56.1} & \hmA{42.7} \\
Gemma-2 27B              & \hmD{78.0} & \hmC{66.8} & \hmB{52.9} & \hmC{69.7} & \hmB{58.4} & \hmA{45.1} \\
Qwen-2.5-32B             & \hmE{82.7} & \hmD{71.2} & \hmB{57.6} & \hmD{74.7} & \hmC{63.4} & \hmA{49.8} \\
Qwen-2.5-72B             & \hmE{85.0} & \hmD{73.9} & \hmB{59.9} & \hmD{77.1} & \hmC{66.0} & \hmB{52.3} \\
LLaMA-3.3-70B            & \hmE{85.4} & \hmD{73.9} & \hmB{59.7} & \hmD{77.6} & \hmC{66.0} & \hmB{52.3} \\
DeepSeek-V2.5            & \hmE{87.3} & \hmD{75.8} & \hmC{62.1} & \hmD{79.6} & \hmC{68.2} & \hmB{54.5} \\
\midrule
PaLM 2                   & \hmE{84.3} & \hmD{72.6} & \hmB{58.8} & \hmD{76.4} & \hmC{65.1} & \hmB{51.4} \\
Claude-3.5 Sonnet        & \hmF{90.1} & \hmD{78.7} & \hmC{65.9} & \hmE{82.3} & \hmD{70.6} & \hmB{56.9} \\
Gemini-2.5-Pro           & \hmE{89.4} & \hmD{78.0} & \hmC{65.1} & \hmE{81.5} & \hmC{69.9} & \hmB{56.1} \\
GPT-4o Mini              & \hmE{87.0} & \hmD{75.5} & \hmC{62.9} & \hmE{80.1} & \hmC{68.8} & \hmB{55.2} \\
GPT-4.1                  & \hmF{91.7} & \hmE{80.3} & \hmD{68.6} & \hmE{84.6} & \hmD{72.8} & \hmB{59.9} \\
GPT-4o                   & \best{\hmF{92.9}} & \best{\hmE{81.8}} & \best{\hmD{70.3}}
                         & \best{\hmE{86.3}} & \best{\hmD{74.6}} & \best{\hmC{62.1}} \\
\bottomrule
\end{tabular}

\vspace{1.0mm}
\centering
\scriptsize
\setlength{\tabcolsep}{3.2pt}
\renewcommand{\arraystretch}{1.05}
\begin{tabular}{lcccccc}
 &
\legbox{blue!6}{\ensuremath{<50}} &
\legbox{blue!14}{\ensuremath{50\mbox{--}<60}} &
\legbox{blue!26}{\ensuremath{60\mbox{--}<70}} &
\legbox{blue!40}{\ensuremath{70\mbox{--}<80}} &
\legbox{blue!56}{\ensuremath{80\mbox{--}<90}} &
\legbox{blue!72}{\ensuremath{\ge 90}} \\
\end{tabular}
\end{table}


\paragraph{Open-Source vs. Proprietary Models.}
While proprietary models outperform open-source systems overall, the gap is narrower for question answering than for summarization. Larger open-source models such as DeepSeek-V2.5 and LLaMA-3.3-70B perform competitively, particularly in English and Arabic, achieving mid-to-high 70s accuracy for question answering. However, summarization remains challenging for all open-source models, suggesting that this task requires stronger discourse-level reasoning and grounding than is currently available in most open-source systems.

\paragraph{Language Effects.}
Performance trends are broadly consistent across languages. English achieves the highest accuracy across models and tasks, followed by Arabic and Turkish, while Hindi consistently exhibits the lowest performance, especially in dialogue summarization. This gap is more pronounced for summarization than for question answering, likely reflecting differences in training data availability and quality for long-form generation in Hindi. Despite these differences, the relative ranking of models remains stable across languages, indicating that hallucination detection capabilities largely transfer across linguistic settings.

\paragraph{Discussion.}
These results highlight the challenges of hallucination detection in multilingual and multi-task scenarios. The persistent performance gap between question answering and dialogue summarization underscores the importance of task-aware evaluation. Moreover, the consistently lower performance on summarization across all languages suggests that detecting subtle, sentence-level hallucinations remains an open problem, even for the strongest available models. \HalluVerse\ thus provides a challenging and realistic benchmark for future work on hallucination detection and mitigation.

\subsection{Per-Hallucination-Type Analysis}
\label{sec:hallucination_type_analysis}

To better understand model behavior beyond aggregate accuracy, we analyze performance separately for each hallucination type: entity-level, relation-level, and sentence-level hallucinations. This analysis is conducted across both tasks and all languages. 
Detailed task- and language-specific results are reported in Tables~\ref{tab:hallucination_en_heatmap}-\ref{tab:hallucination_tr_heatmap}.

\paragraph{Entity-Level Hallucinations.}
Entity hallucinations are consistently the easiest category for models to detect across tasks and languages. As shown in Tables~\ref{tab:hallucination_en_heatmap}--\ref{tab:hallucination_tr_heatmap}, frontier models such as GPT-4o and GPT-4.1 achieve very high accuracy for question answering, exceeding \kval{90\%} in all languages and reaching above \kval{95\%} in English. Strong open-source models, including DeepSeek-V2.5 and LLaMA-3.3-70B, also perform competitively, typically achieving accuracy in the mid-to-high 80s for question answering. Performance on entity-level hallucinations decreases for dialogue summarization across all models and languages, with drops ranging from approximately \kval{6 to 10} percentage points. This decline reflects the increased difficulty of identifying incorrect entities when they appear within longer and less structured summaries rather than as localized factual errors.

\paragraph{Relation-Level Hallucinations.}
Relation hallucinations are more challenging than entity hallucinations across all settings. These hallucinations involve incorrect temporal, causal, or attributional relationships, which require deeper semantic reasoning to detect. Across languages, question answering accuracy for relation hallucinations generally falls in the \kval{70--85\%} range for frontier models and in the \kval{60--75\%} range for strong open-source models. In dialogue summarization, performance consistently declines, as relational errors are often embedded within broader contextual narratives, making them harder to isolate.

\paragraph{Sentence-Level Hallucinations.}
Sentence-level hallucinations are the most difficult category across both tasks and all languages. Even the strongest models achieve only moderate performance in this setting. Accuracy typically ranges between \kval{65--75\%} for question answering and drops to approximately \kval{50--65\%} for dialogue summarization. These hallucinations often introduce fluent but unsupported content that is not anchored to specific entities or relations, reducing the effectiveness of surface-level detection cues.

\paragraph{Task-Specific Trends.}
Across all hallucination types, question answering consistently yields higher performance than dialogue summarization. The performance gap is smallest for entity-level hallucinations and largest for sentence-level hallucinations. This pattern suggests that hallucinations spanning longer textual segments, which are common in summarization, pose a substantially greater challenge than localized factual errors typically observed in question answering.

\paragraph{Language Effects.}
While overall performance varies across languages, several consistent trends emerge. English yields the strongest results across hallucination types and tasks, followed by Arabic and Turkish, with Hindi consistently presenting the greatest challenge. Importantly, the relative ordering of hallucination difficulty remains stable across languages: entity-level hallucinations are easiest to detect, followed by relation-level hallucinations, with sentence-level hallucinations being the hardest. This consistency indicates that the observed trends are driven primarily by hallucination structure and task characteristics rather than language-specific artifacts.

\paragraph{Summary.}
Overall, the per-hallucination-type analysis reveals clear and consistent performance gradients across models, tasks, and languages. While entity-level hallucinations are largely tractable for current frontier models, relation- and sentence-level hallucinations remain substantially more challenging, particularly in generative tasks such as dialogue summarization. These findings underscore the importance of fine-grained evaluation and motivate the need for benchmarks that explicitly distinguish between hallucination types.

\section{Conclusion}
\label{sec:conclusion}
We introduced \HalluVerse,\ a multilingual and multi-task dataset designed to support fine-grained analysis of hallucinations in natural language generation. The dataset spans question answering and dialogue summarization, covers four languages, and explicitly distinguishes between entity-level, relation-level, and sentence-level hallucinations. Through a controlled construction pipeline and human validation, \HalluVerse\ provides high-quality supervision for studying hallucination behavior across diverse settings.  Our empirical evaluation demonstrates that hallucination detection remains a challenging problem, particularly for sentence-level hallucinations and abstractive generation tasks such as summarization. While recent frontier models achieve strong performance in question answering, substantial performance gaps persist in summarization and across languages. These findings highlight the limitations of current models in detecting subtle, context-dependent hallucinations and underscore the importance of task-aware and type-aware evaluation. By enabling systematic comparisons across tasks, languages, and hallucination types, \HalluVerse\ offers a valuable benchmark for future research on hallucination detection and mitigation. We hope this dataset will facilitate more nuanced evaluations of generative models and encourage the development of methods that move beyond surface-level cues toward deeper semantic grounding.

\section*{Ethical Statement}

\paragraph{Intended Use and Potential Misuse.}
\HalluVerse\ is designed to support research on hallucination detection and analysis in multilingual and multi-task generation settings. The dataset contains deliberately modified outputs that include incorrect or fabricated information. As such, it could be misused to train or evaluate systems that generate misleading content if applied outside its intended research scope. There is also a risk that hallucinated outputs could be taken out of context and redistributed as factual statements. To reduce these risks, all hallucinated instances are explicitly labeled and paired with their corresponding original outputs, making the presence and type of hallucination clear. The dataset is released for non-commercial research use only under a \textbf{Creative Commons Attribution--NonCommercial 4.0 International (CC BY-NC 4.0)} license, and access is gated to ensure that users acknowledge and agree to the stated terms of use.

\paragraph{Content Safety and Harmful Language.}
The dataset construction process prioritizes factual tasks and controlled generation, which substantially limits the introduction of abusive or offensive language. To further assess potential risks, we performed automated screening on the English portion of the dataset using publicly available offensive language detection models from Hugging Face, including \textsc{KoalaAI/OffensiveSpeechDetector}\footnote{\href{https://huggingface.co/KoalaAI/OffensiveSpeechDetector}{Model card on Hugging Face}} and \textsc{distilroberta-base-offensive}\footnote{\href{https://huggingface.co/badmatr11x/distilroberta-base-offensive-hateful-speech-text-multiclassification}{Model card on Hugging Face}}.
These analyses indicated a low presence of offensive content (\kval{0.098} and \kval{0.019}, respectively). While equivalent tools are not consistently available for all languages covered in the dataset, the use of task-grounded source material and subsequent human verification further reduces the likelihood of harmful language. We consider content safety to be an ongoing responsibility and encourage users to report any issues that may arise.


\bibliographystyle{named}


\appendix

\end{document}